\pdfoutput=1

\documentclass[11pt]{article}

\usepackage{acl}

\usepackage{times}
\usepackage{latexsym}

\usepackage[T1]{fontenc}

\usepackage[utf8]{inputenc}

\usepackage{microtype}

\usepackage{tabularx}
\usepackage{multicol}
\usepackage{multirow}

\usepackage{graphicx}
\usepackage{subfig}

\usepackage{hyperref}

\usepackage[ruled,vlined]{algorithm2e}

\usepackage{xcolor,colortbl}
\definecolor{darkgreen}{HTML}{008A22}
\definecolor{green}{HTML}{76d275}
\definecolor{yellow}{HTML}{b2fab4}
\definecolor{blue}{HTML}{d7ffd9}

\usepackage{booktabs}

\title{Use of a Taxonomy of Empathetic Response Intents to Control and Interpret Empathy in Neural Chatbots}

\author{Anuradha Welivita, and Pearl Pu\\
  School of Computer and Communication Sciences \\
  École Polytechnique Fédérale de Lausanne \\
  Switzerland \\
  \texttt{\{kalpani.welivita,pearl.pu\}@epfl.ch}}

\begin{document}
\maketitle
\begin{abstract}


A recent trend in the domain of open-domain conversational agents is enabling them to converse empathetically to emotional prompts. Current approaches either follow an end-to-end approach or condition the responses on similar emotion labels to generate empathetic responses. But empathy is a broad concept that refers to the cognitive and emotional reactions of an individual to the observed experiences of another and it is more complex than mere mimicry of emotion. Hence, it requires identifying complex human conversational strategies and dynamics in addition to generic emotions to control and interpret empathetic responding capabilities of chatbots. In this work, we make use of a taxonomy of eight empathetic response intents in addition to generic emotion categories in building a dialogue response generation model capable of generating empathetic responses in a controllable and interpretable manner. It consists of two modules: 1) a response emotion/intent prediction module; and 2) a response generation module. We propose several rule-based and neural approaches to predict the next response's emotion/intent and generate responses conditioned on these predicted emotions/intents. Automatic and human evaluation results emphasize the importance of the use of the taxonomy of empathetic response intents in producing more diverse and empathetically more appropriate responses than end-to-end models.
\end{abstract}


\section{Introduction}

End-to-end neural dialogue response generation has revolutionized the design of open-domain conversational agents or chatbots due to requiring little or no manual intervention and its ability to largely generalize \cite{sordoni,shang, vinyals}. It overcomes many limitations of traditional rule-based response generation techniques such as the cost of domain expertise and predictability of responses. But due to the black-box nature of these end-to-end models, they offer very little controllability to the developer and generate responses that are difficult to interpret \cite{microsoft, microsoft2, cmu}, making these approaches less reliable and fail-safe \cite{Neurosymbolic}. A recent example is Microsoft's Taybot that started producing unintended, and offensive tweets denying the Holocaust as a result of learning from racist and offensive information on Twitter \cite{lee_2016}. Having control over the generated responses would have enabled the chatbot to avoid malicious intentions and carefully choose how to converse. Thus, it is important to look at ways how developers can gain control over the responses generated by end-to-end neural response generation models and how they can be made interpretable.



Recent research has taken efforts to induce controllability and interpretability into end-to-end models. For example, Xu et al. \shortcite{microsoft} explore how the flow of human-machine interactions can be managed by introducing dialogue acts as policies to the dialogue generation model. Sankar and Ravi \shortcite{google} show that conditioning the response generation process on interpretable dialogue attributes such as dialogue acts and sentiment helps to eliminate repetitive responses and makes the model more interesting and engaging.



\begin{table}
\scriptsize
\centering
\begin{tabularx}{\linewidth}{r X}
\toprule
\multicolumn{2}{l}{\textbf{Dialogue context:}}\vspace{1mm}\\

Speaker: & \textit{I think that the girl of my dreams likes somebody else. I feel very sad about it.}\\
Listener: & \textit{Ooh, am so sorry about that. \textcolor{red}{Have you tried to talk to her?}}\\
Speaker: & \textit{It's tough as she has been out of the country for a month, so I will likely discuss it when she returns.}\\

\midrule

\multicolumn{2}{l}{\textbf{Possible responses:}}\vspace{1mm}\\
(No control) & \textit{\textcolor{red}{Have you talked to her about it yet?}} (Repetitive)\\
(No control) & \textit{\textcolor{red}{I don't think that's a good idea.}} \\
& (Not encouraging to the speaker)\\
  
 (Conditioned on: & \textit{\textcolor{darkgreen}{I hope everything works out for you.}} \\
 \textbf{Encouraging}) & (Empathetically appropriate) \\

\bottomrule
\end{tabularx}
\caption{An example dialogue showing how controllability affects response generation.}
\label{tab:example-init}
\end{table}



In contrast to task-oriented dialogue systems designed to help people complete specific tasks, open-domain chatbots are designed to engage users in human-machine conversation for entertainment and emotional companionship \cite{wu_and_yan}. Hence, in open-domain conversations, controllability should also be studied with respect to aspects such as humor, personality, emotions, and empathy, which cannot be achieved using generic dialogue acts. In this study, our focus is on controlling empathy in open-domain chatbot responses, which requires understanding conversational strategies used in human-human empathetic conversations. 

Earlier studies gain control in this aspect by conditioning the response on either manually specified \cite{ecm,mojitalk,Hu,Song} or automatically predicted \cite{Chen} sentiment or emotion labels. However, an analysis by Welivita and Pu \shortcite{taxonomy} on human-human conversations of the EmpatheticDialogues dataset \cite{empatheticdialogues} reveals, listeners are much more likely to respond to positive or negative emotions with specific empathetic intents such as \textit{acknowledgment}, \textit{consolation} and \textit{encouragement}, rather than expressing similar or opposite emotions. They introduce a taxonomy of eight response intents that can better describe empathetic human responses to emotional dialogue prompts. In this paper, we explore how end-to-end response generation can be combined with more advanced control of empathy by utilizing the above taxonomy of empathetic response intents in addition to existing emotion categories. To provide a glimpse of what we aim to achieve, in Table \ref{tab:example-init} we show how conditioning the response on an empathetic response intent chosen based on the dialogue history can serve in producing a more empathetically appropriate response. It avoids repetitive or sub-optimal responses generated by end-to-end approaches without any control.


Our empathetic response generation model consists of two modules: 1) a response emotion or intent prediction module; and 2) a response generation module. We experiment with both rule-based and neural approaches for predicting the next response's emotion or intent. For the rule-based approaches for predicting the response emotion/intent, we develop two decision tree-based response emotion and intent prediction methods. For the neural approach for predicting the response emotion/intent, we develop a classifier based on the BERT transformer architecture \cite{transformer, bert}. The reason why we evaluate the performance of rule-based approaches is that they are much simpler than neural models and save a lot of training time and resources. Thus, if considerable performance can still be achieved through rule-based approaches compared to the baselines, it is worth considering the use of such simpler approaches over sophisticated neural approaches, especially in resource-limited environments. The emotions and intents predicted by these methods are then used to condition the responses generated by the response generation module. For training and evaluating these models, we use two state-of-the-art dialogue datasets containing empathetic conversations: 1) the EmpatheticDialogues dataset \cite{empatheticdialogues}; and 2) the EDOS (Emotional Dialogues in OpenSubtitles) dataset \cite{edos}. The automatic and human evaluation results confirm the importance of the use of the taxonomy in generating more diverse and empathetically more appropriate responses than end-to-end models.



Our contributions in this paper are three folds. 1) We explore the ability of a taxonomy of empathetic response intents in controlling and interpreting the responses generated by open-domain conversational agents for emotional prompts. 2) We propose an empathetic response generation model consisting of a response emotion/intent prediction module and a response generation module to generate empathetic responses in a controllable and interpretable manner. 3) We experiment with both rule-based and neural approaches in predicting the next response's emotion or intent and evaluate their performance in conditional generation of empathetic responses using automatic and human evaluation metrics against standard baselines.

\section{Literature Review}

Existing conversational agents are designed for either open-domain or specific task completion \cite{survey}. Regarding the former, a common practice is to generate dialogue in an end-to-end fashion \cite{sordoni,shang, vinyals}. Often responses generated by these methods are unpredictable and not fail-safe \cite{Neurosymbolic}. Hence, recent research has focused on methods to control and interpret the responses generated by open-domain neural conversational agents. Mainly we find three methods they use to control the generated response: 1) by a manually specified value \cite{ecm, mojitalk, Hu, Song}; 3) by rules that are predefined or derived from the training data \cite{alexa}; 3) by an automatically predicted value from a neural network model \cite{meed2, microsoft, google, north_carolina, tsinghua, ntu}.

Specially in studies addressing emotional response generation, a manually specified sentiment, emotion \cite{ecm} or an emoji \cite{mojitalk} was used to control the sentiment or emotionality of the responses generated. Later, more and more research focused on automatically predicting values or deriving rules such that they could be used to control the generated response without manual intervention. For example, Sankar and Ravi \shortcite{google} used an RNN based policy network to predict the next dialogue act given previous dialogue turns and dialogue attributes. Hedayatnia et al. \shortcite{alexa} used rules designed as a set of dialogue act transitions from common examples in the Topical-Chat corpus \cite{topical_chat} to plan the content and style of target responses.

But all the above work focused on achieving controllability using generic dialogue acts or generating controlled emotional responses conditioned on similar or opposite emotions, emojis, or sentiment tags. These labels do not suffice the controlled generation of meaningful empathetic responses because humans demonstrate a wide range of emotions and intents when regulating empathy \cite{taxonomy}. Previous work also lacks comparisons between rule-based and automatic conditioning methods used to control response generation. In this work, we address the above gaps by investigating how empathy in neural responses can be controlled using a taxonomy of eight empathetic response intents \cite{taxonomy}, in addition to 32 emotion categories, while evaluating the applicability of both rule-based and automatic control mechanisms for this task. 

\section{Methodology}


Our controllable and interpretable empathetic response generation architecture consists of two modules: 1) the response emotion/intent prediction module; and 2) the response generation module. The emotion or intent predicted by the first module is input into the second to condition the response generated by the second module. In the following sections we discuss the datasets used for our experiments, the different rule-based and automatic emotion/intent prediction methods we propose, how the emotions and intents predicted by these modules are used to generate responses that are both controllable and interpretable, and the different evaluation methods we utilize to compare the performance of these approaches on two state-of-the-art dialogue datasets containing emotional dialogue prompts.

\subsection{Datasets}

We utilized the EmpatheticDialogues dataset proposed by Rashkin et al. \shortcite{empatheticdialogues}, and the OS (OpenSubtitles) and EDOS (Emotional Dialogues in OpenSubtitles) dialogue datasets proposed by Welivita et al. \shortcite{edos} to train and evaluate our models. The EmpatheticDialogues (ED) dataset contains $\approx$25K open-domain human-human conversations carried out between a speaker and a listener. Each conversation is conditioned on one of 32 emotions selected from multiple annotation schemes. The OS and EDOS datasets are curated by applying a series of preprocessing and turn segmentation steps on the movie and TV subtitles in the OpenSubtitles 2018 corpus \cite{opensubtitles}. The EDOS dataset contains 1M highly emotional dialogues filtered from the rest of the OS dialogues. Even though the speaker and listener turns in the OS and EDOS datasets are not clearly defined, we assumed the odd-numbered turns (1, 3, 5, ...) as speaker turns and even-numbered turns (2, 4, 6, ...) as listener turns for our experiments. We used the OS dialogues dataset containing $\approx$3M dialogues for pre-training and the ED and EDOS datasets to separately fine-tune the models. The statistics of these datasets are denoted in Table \ref{tab:datasets}. From each dataset, 80\% of the data was used for training, 10\% for validation, and the remaining 10\% for testing. 

\begin{table}[ht!]
\scriptsize
\centering
\begin{tabular}{l r r r}
\toprule
Dataset  & Dialogues & Turns & Turns/dialogue\\
\midrule
OS\vspace{1mm} & 2,989,774 & 11,511,060 & 3.85\\
ED\vspace{1mm} & 24,847 & 107,217 & 4.32\\
EDOS & 1,000,000 & 2,940,629 & 2.94\\

\bottomrule
\end{tabular}
\caption{Statistics of the datasets used for training and evaluating the models.}
\label{tab:datasets}
\end{table}

We used a BERT \cite{bert} transformer-based dialogue emotion classifier proposed by Welivita et al. \shortcite{emobert+} to automatically annotate all dialogue turns in the above datasets. This classifier is trained on 25K situation
descriptions from EmpatheticDialogues labeled with 32 emotion classes, 7K EmpatheticDialogues listener turns labeled with eight empathetic response intents and \textit{Neutral}, and 14K emotion and intent annotated dialogue turns from the OSED dataset. It has a final annotation accuracy of 65.88\% over 41 labels, which is significant compared to the other state-of-the-art dialogue emotion classifiers \cite{emobert+}. We use the emotion and intent labels suggested by the above classifier as ground-truth labels for our experiments. 


\subsection{Response Emotion/Intent Prediction}

To generate controlled and interpretable empathetic responses, we utilized 32 fine-grained emotions and a taxonomy of listener-specific empathetic response intents. The 32 emotions are emotion categories on which dialogues in the EmpatheticDialogues dataset are conditioned on \cite{empatheticdialogues}. They range from basic emotions derived from biological responses \cite{ekman,plutchik} to larger sets of subtle emotions derived from contextual situations \cite{skerry}. We further utilized the taxonomy of empathetic response intents proposed by Welivita and Pu \shortcite{taxonomy}, which is derived by analysing the listener responses in the EmpatheticDialogues dataset. These intents are denoted in Table \ref{tab:intents} along with corresponding examples. To predict the emotion or intent of the next response, we propose several rule-based and neural response emotion/intent prediction methods, which are described in the following subsections.  

\begin{table}
\scriptsize
\centering
\begin{tabular}{l l}
\toprule
Empathetic intent  & Example response\\
\midrule

1. Questioning & \textit{What's the matter? What's wrong?}\\
2. Agreeing & \textit{Exactly, I get that entirely!}\\
3. Acknowledging  & \textit{Sounds awesome!}\\
4. Encouraging & \textit{Just give it a trial.}\\
5. Consoling & \textit{I hope everything works out for you.}\\
6. Sympathizing & \textit{I am sorry to hear that.}\\
7. Wishing & \textit{Congrats, that's a step forward.}\\
8. Suggesting & \textit{Maybe you should talk to her.}\\


\bottomrule
\end{tabular}
\caption{The taxonomy of listener specific empathetic response intents used to achieve controllability and interpretability in the responses generated.}
\label{tab:intents}
\end{table}

\subsubsection{Baselines}

As a baseline, we sample a response emotion or intent from the set of eight empathetic response intents plus the most recent emotion encountered in the last $k (k=3)$ dialogue turns. This is based on the observations by Welivita and Pu \shortcite{taxonomy} on the EmpatheticDialogues dataset \cite{empatheticdialogues}. They state that in human empathetic conversations, the listener's response to emotional prompts mostly contain an empathetic response intent identified by their taxonomy or a statement with similar emotion. This baseline is inspired by the work of Hedayatnia et al. \shortcite{alexa}, in which the response dialogue act is chosen among the most frequently seen dialogue acts based on an equal probability distribution.  

The other baseline we used when generating responses is the plain end-to-end transformer model proposed by Vinyals et al. \shortcite{transformer}, in which no conditioning is used when generating the response.

\subsubsection{Rule-based Decision Tree Approaches}

\begin{figure*}[ht!]
\centering
\includegraphics[width=0.8\linewidth]{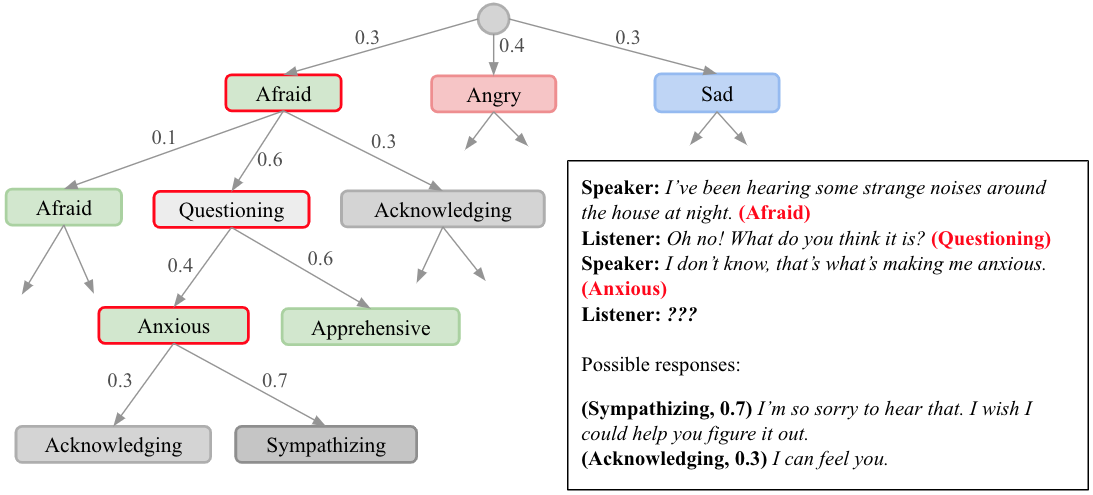} 
\caption{Visualization of a simpler version of our decision tree approach to predict the response emotion or intent.}
\label{fig:decision_tree}
\end{figure*}

\begin{figure*}[ht!]%
    \centering
    
    \subfloat[\centering Dataset: EmpatheticDialogues]{{\includegraphics[width=0.46\linewidth]{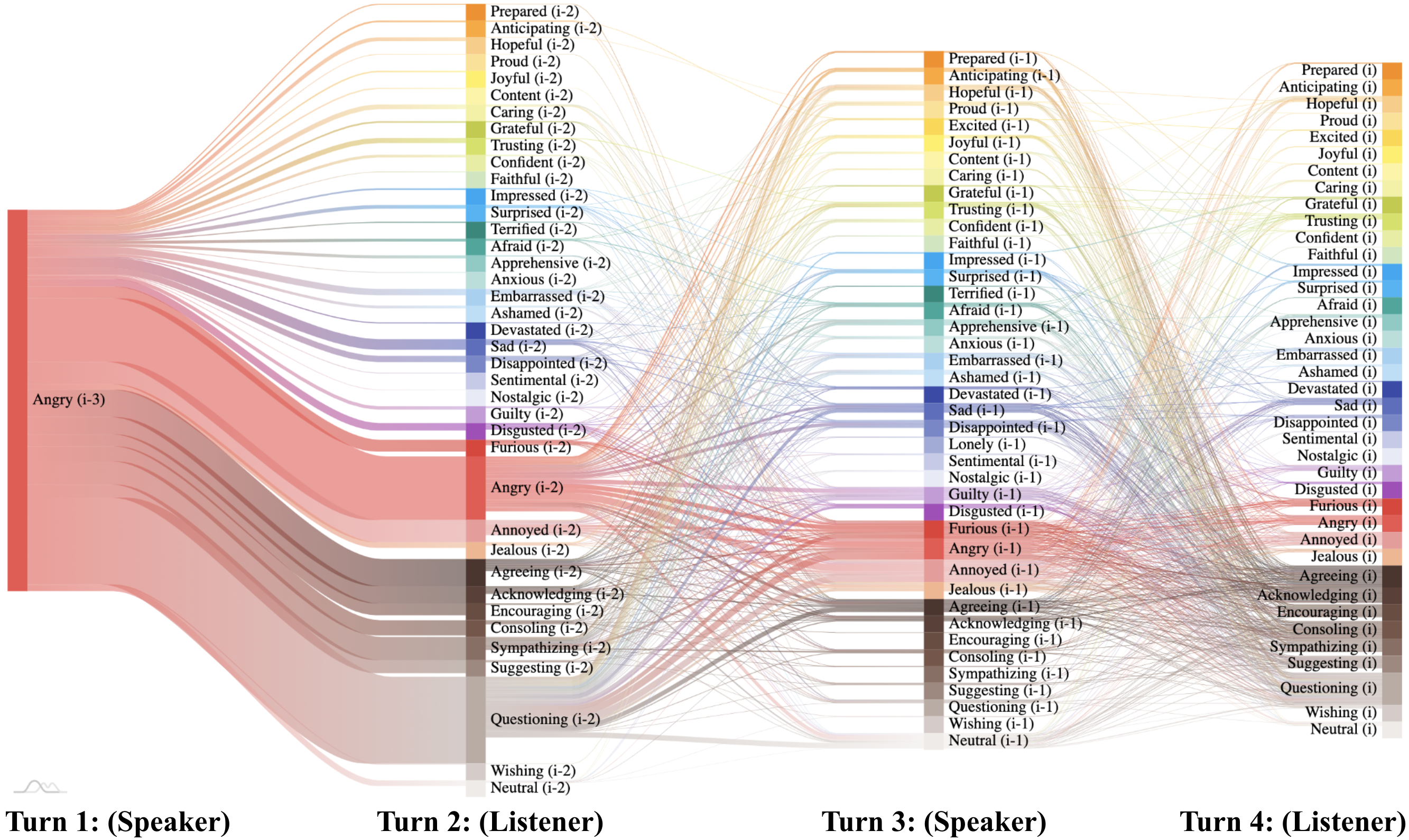} }}%
    \qquad
    \subfloat[\centering Dataset: EDOS]{{\includegraphics[width=0.46\linewidth]{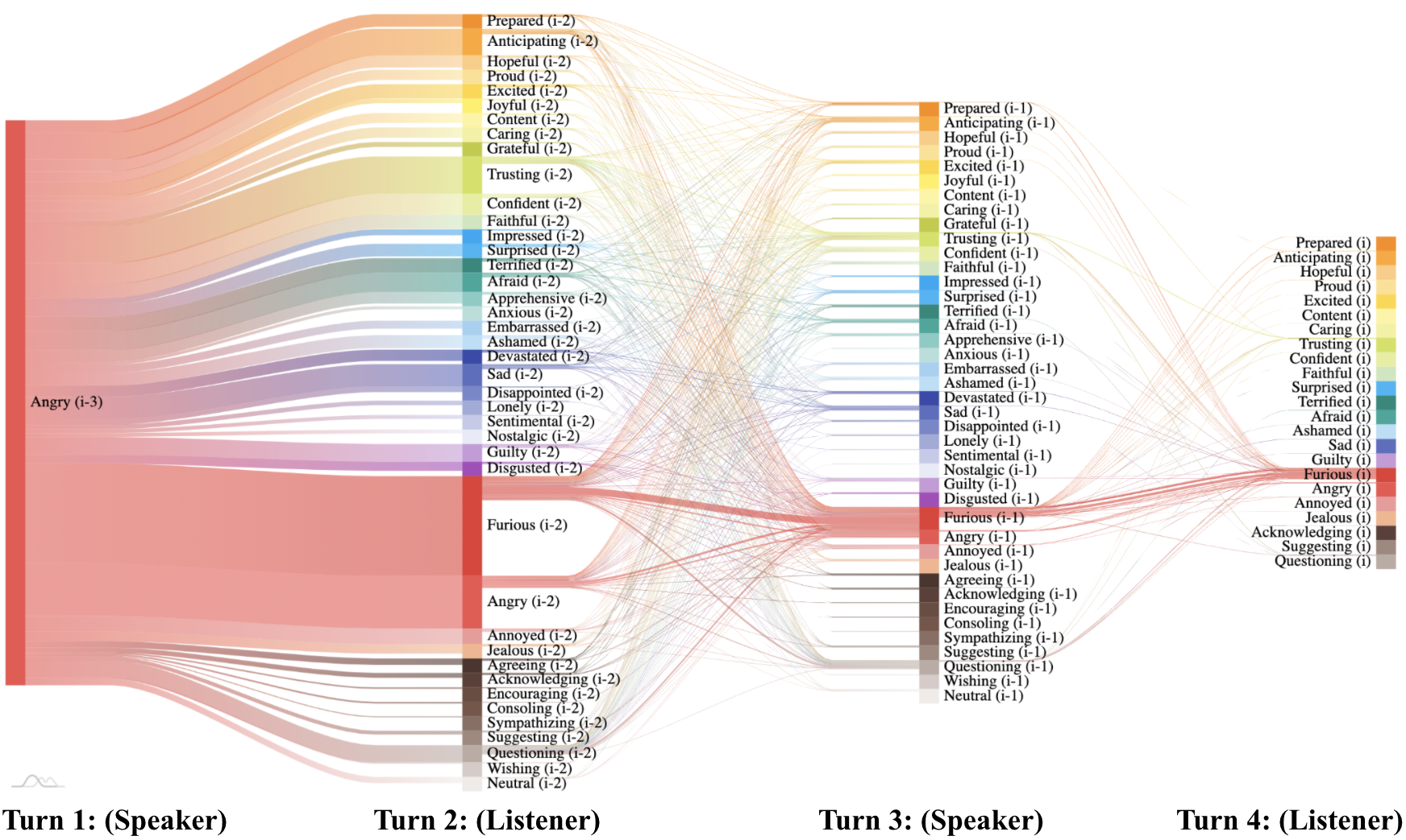} }}%


    \caption{Decision trees generated using the EmpatheticDialogues and EDOS training datasets when the emotion of the beginning dialogue prompt is \textit{Angry}.}%
    \label{fig:decision_tree_visualization}%
\end{figure*}

We propose two non-neural, decision tree-based response intent prediction methods that leverage the knowledge of the emotion-intent flow of the dialogues in the training dataset. The basic idea of a decision tree for this context is denoted along with an example in Figure \ref{fig:decision_tree}. The probabilities of emotions and intents in the branches in the decision tree are learned from the training data itself by traversing through dialogues using a window of size $k$, where $k$ is the maximum depth of the decision tree. The window is moved forward two dialogue turns at a time capturing the probability of speaker-listener emotion-intent exchanges in the training dataset. 


Here, we used a window of size $4$ mainly because most dialogues contained in the ED, OS, and EDOS datasets were limited to four dialogue turns. During inference, an emotion or an intent is sampled based on the sequence of emotions and intents in the previous $(k-1)$ dialogue turns. We used two different methods: 1) argmax; and 2) probabilistic sampling, to sample the response emotion or intent from the decision tree. In the argmax method, we chose the emotion or intent with the highest probability in the decision tree based on the sequence of emotions and intents in the previous $(k-1)$ dialogue turns. In the probabilistic sampling method, we sampled an emotion or an intent based on the distribution of probabilities in the decision tree given the sequence of emotions and intents in the previous $(k-1)$ dialogue turns. We refer to these two decision tree-based methods as \textit{DT (argmax)}, and \textit{DT (prob. sampled)}.



We have more control over the above methods than neural response intent prediction methods since we can foresee where the dialogue will be directed by visualizing the decision trees beforehand. For example, the decision trees generated using the EmpatheticDialogues and EDOS training datasets when the emotion of the beginning dialogue prompts is \textit{Angry} are denoted in Figure \ref{fig:decision_tree_visualization}. As it could be observed, in the ED dataset, the listeners mostly respond to speakers' emotions with one of the intents from the taxonomy of empathetic response intents. The EDOS dataset by nature is more dramatic, in which both the speaker and the listener become emotional. This phenomenon is called ``emotional contagion" in the psychological literature \cite{emotional}. For example in EDOS, if the speaker is angry, the listener also tends to reply back with anger. These communication patterns could clearly be visualized with the decision trees created and the developer can predict beforehand how the chatbots whose responses are conditioned on these emotion-intent patterns would behave for a given emotional prompt. 

\subsubsection{Neural Response Emotion and Intent Predictor}
\label{sec:neural_pred}

\begin{figure}[ht!]
\centering
\includegraphics[width=\linewidth]{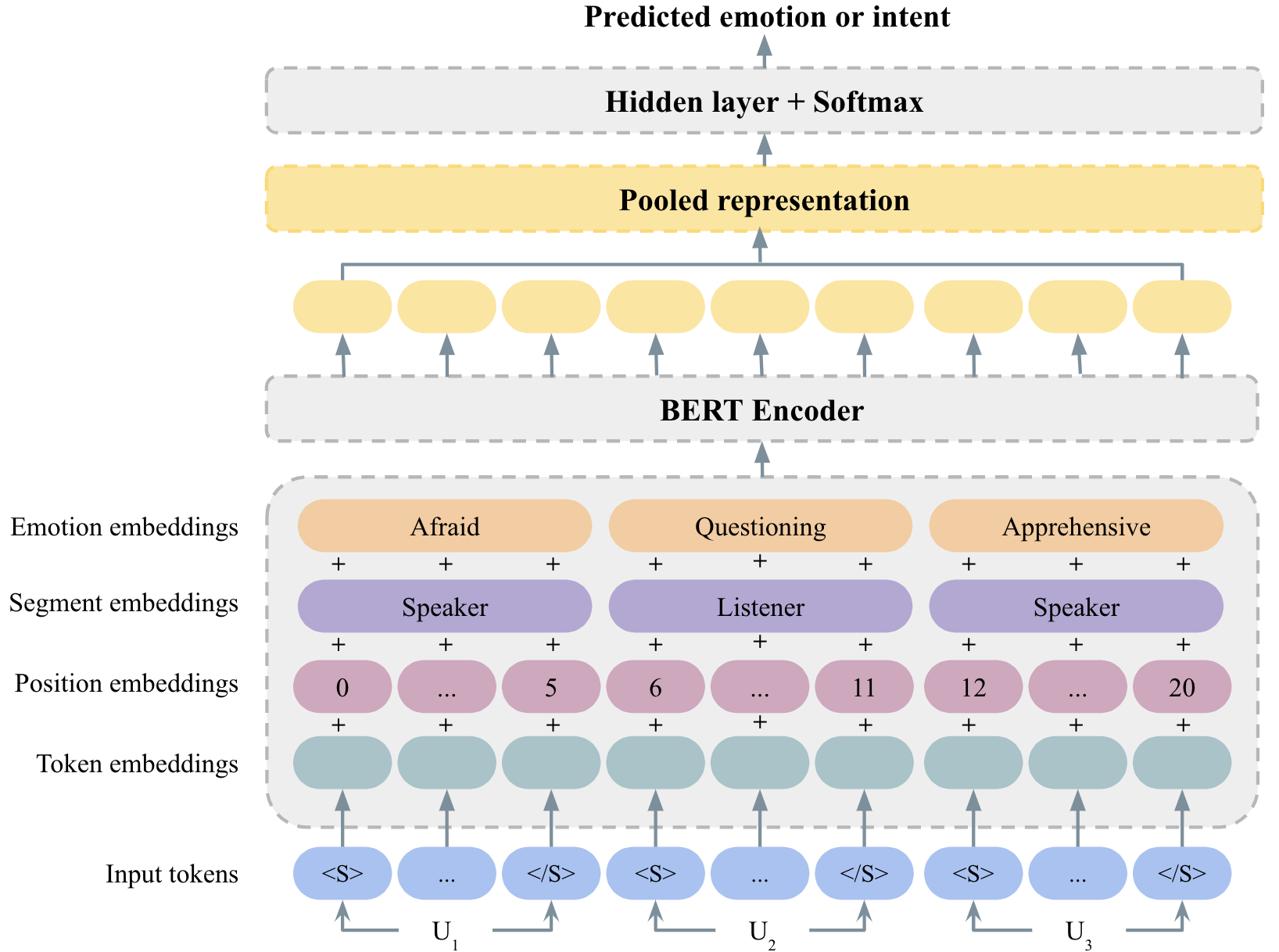} 
\caption{Architecture of the neural response emotion/intent predictor.}
\label{fig:meed2}
\end{figure}

An automatic method for predicting the next response's emotion or intent is using a neural network-based response emotion/intent predictor. An advantage of using neural approaches to determine the emotion or intent of the next response is that they can leverage clues from the semantic content of the previous dialogue turns in addition to the flow of emotions and intents when predicting the response emotion or intent. Our neural response emotion/intent predictor consists of a BERT transformer-based encoder architecture (representation network) followed by an attention layer for aggregating individual token representations, a hidden layer, and a softmax as depicted in Figure \ref{fig:meed2}. The BERT-base architecture with 12 layers, 768 dimensions, 12 heads, and 110M parameters is used as the representation network. It is initialized with weights from the pre-trained language model RoBERTa \cite{roberta}.


We concatenate the previous $k$ dialogue turns as depicted in Figure \ref{fig:meed2} and they are input to the encoder of the model. The emotions and intents corresponding to these $k$ dialogue turns are added to the word embeddings and positional embeddings in the original transformer architecture. This additional knowledge helps the model to get a better understanding of the flow of emotions and intents in the previous dialogue turns. The emotions and intents are embedded into a vector space having the same dimensionality as the word and position embeddings so they can add up. In addition, we also incorporate segment embeddings that differentiate between speaker and listener turns. We pre-trained the model on the OS dialogues dataset and fine-tuned it separately on ED and EDOS datasets. The hyper-parameters used during training and other training details are described in the appendices.

\subsection{Response Generation}

\begin{figure}[ht!]
\centering
\includegraphics[width=\linewidth]{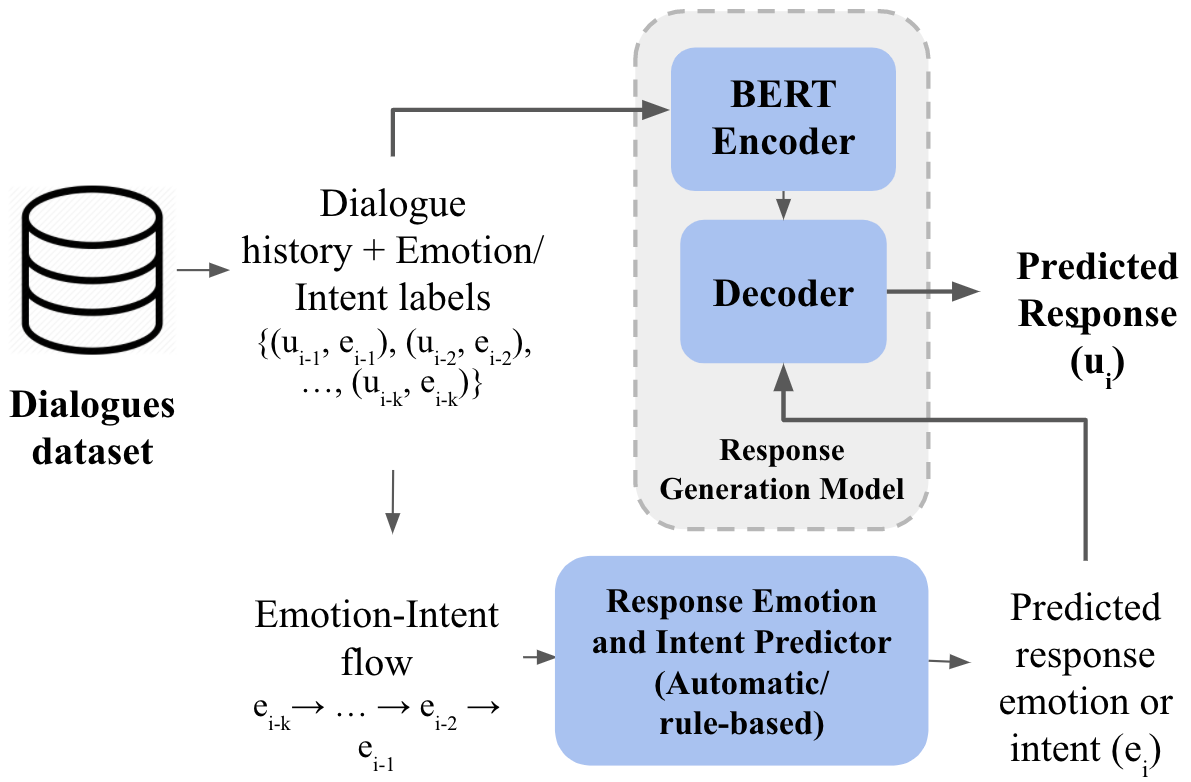} 
\caption{Overall architecture of the controllable and interpretable empathetic response generation model.}
\label{fig:overall_architecture}
\end{figure}

For response generation, we used a plain transformer-based encoder-decoder architecture (end-to-end model) as a baseline \cite{transformer}. To generate controlled empathetic responses, we incorporated the different response emotion/intent prediction methods described above as input to the decoder. Figure \ref{fig:overall_architecture} shows the overall architecture of our models. 


The input representation for the encoder of the generation model is the same as the input representation used for the neural response emotion/intent predictor described in section \ref{sec:neural_pred}. The vector representation generated by the encoder is input into the decoder along with the embedding of the emotion or intent predicted by the response emotion/intent predictor. During training, instead of the predicted emotion or intent, we used the ground-truth emotion or intent. The generation model is first pre-trained on OS dialogues and then fine-tuned on ED and EDOS datasets separately.

\begin{table*}
\centering
\scriptsize
\begin{tabular}{l | r r r r | r r r r }
\toprule

\multirow{3}{*}{Model} & \multicolumn{4}{c|}{Trained on: OS + ED} & \multicolumn{4}{c}{Trained on: OS + EDOS} \\
& \multicolumn{4}{c|}{Tested on: ED} & \multicolumn{4}{c}{Tested on: EDOS} \\
\cline{2-9}
& Prec. & Recall & F1 & Acc. & Prec. & Recall & F1 & Acc.\\

\hline


Equally sampled & \cellcolor{yellow} 0.1138	& 0.0667	& 0.0638	& 0.0410 & \cellcolor{yellow} 0.0981 & 0.0221 & 0.0232 & 0.0285\\
DT (argmax) &  0.0959	& \cellcolor{yellow} 0.0883	& \cellcolor{yellow} 0.0883	& \cellcolor{yellow} 0.0692 & 0.0755 & \cellcolor{yellow} 0.1016 & \cellcolor{yellow} 0.0799 & \cellcolor{yellow} 0.0419\\
DT (prob. sampled) & 0.0715	& 0.0663	& 0.0680	& 0.0480 & 0.0627 & 0.0616 & 0.0619 & 0.0345\\
Neural predictor & \cellcolor{green} 0.1634	& \cellcolor{green} 0.1636	& \cellcolor{green} 0.1472	& \cellcolor{green} 0.1163 & \cellcolor{green} 0.1306 & \cellcolor{green} 0.1712 & \cellcolor{green} 0.1181 & \cellcolor{green} 0.0679\\

\bottomrule
\end{tabular}
\caption{Weighted precision, recall, F1 and accuracy scores computed for ED and EDOS test datasets. The cells in dark green indicate the best scores and the cells in light green indicate the second best scores.}
\label{tab:pred_scores}
\end{table*}

\begin{table*}
\centering
\scriptsize
\begin{tabular}{ l | r r r r | r r r r r}
\toprule

\multirow{3}{*}{Model} & \multicolumn{4}{c|}{Trained on: OS + ED} & \multicolumn{4}{c}{Trained on: OS + EDOS} \\
& \multicolumn{4}{c|}{Tested on: ED} & \multicolumn{4}{c}{Tested on: EDOS} \\
\cline{2-9}
 & PPL & D-1 & D-2 & Embed. & PPL & D-1 & D-2 & Embed.\\
& & & & extrema & & & & extrema\\

\hline

GT emotion/intent & \cellcolor{green} 11.74 & 0.0823 & \cellcolor{yellow} 0.2812 & \cellcolor{green} 0.5181 & \cellcolor{green} 12.57 & \cellcolor{green} 0.0846 & \cellcolor{green} 0.2552 & \cellcolor{green} 0.4539\\

End-to-end model & \cellcolor{yellow} 12.26 & 0.0544 & 0.1612 & \cellcolor{yellow} 0.5015 & \cellcolor{yellow} 13.13 & 0.0784 & 0.228 & 0.4365\\

Equally sampled & 13.48 & 0.0761 & 0.2469 & 0.4824 & 14.20 & 0.0754 & 0.2229 &  0.433\\

 
DT (argmax) & 13.23 & \cellcolor{green} 0.0865 & \cellcolor{green} 0.2977 & 0.4892 & 14.14 & 0.0727 & \cellcolor{yellow} 0.2419 & \cellcolor{yellow} 0.4458\\
 
 
DT (prob. sampled) & 13.37 & 0.0795 & 0.2761 &  0.4828 & 14.23 & 0.0763 & 0.2418 &  0.436\\
 
Neural predictor & \cellcolor{blue} 13.15 & \cellcolor{yellow} 0.0835 & 0.2811 & 0.4851 & \cellcolor{blue} 13.97 & \cellcolor{yellow} 0.0805 & 0.2415 & 0.4403\\

\bottomrule
\end{tabular}
\caption{Perplexity (PPL), diversity metrics (distinct unigrams: D-1; and distinct bigrams: D-2), and vector extrema cosine similarity (Embed. extrema) calculated on ED and EDOS testing datasets.}
\label{tab:res_scores}
\end{table*}

\section{Evaluation and Results}


\subsection{Automatic Evaluation Results}

Evaluation by means of automatic metrics was carried out separately for response emotion/intent prediction and conditional response generation. The following subsections describe the results obtained in these evaluations.  

\subsubsection{Prediction Performance}

The weighted precision, recall, F1, and balanced accuracy scores computed for different response emotion/intent prediction methods across ED and EDOS testing datasets are indicated in Table \ref{tab:pred_scores}. 


According to the weighted precision, recall, F1, and accuracy scores, the neural emotion/intent predictor performed the best compared to other prediction methods. Among rule-based approaches for response emotion/intent prediction, the DT (argmax) method performed the best. The DT (argmax) method had considerable improvement in recall, F1, and accuracy scores over the equally sampled baseline. 


\subsubsection{Generation Performance}

To evaluate the performance of response generation, we computed the perplexity, diversity metrics (distinct unigram and distinct bigram scores), and vector extrema cosine similarity on ED and EDOS testing datasets. They are denoted in Table \ref{tab:res_scores}. We also evaluated the responses generated by a model conditioned on the ground-truth emotion or intent of the next response to see how well the taxonomy of empathetic response intents alone contributes to better empathetic response generation performance. 

According to the results, the models whose response was conditioned on the ground-truth response emotion or intent performed the best in terms of perplexity and embedding extrema in both ED and EDOS datasets and in terms of diversity metrics in the EDOS dataset. These results emphasize the usefulness of the taxonomy of empathetic response intents and the 32 fine-grained emotion categories in generating controlled empathetic responses. The models incorporating the DT (argmax) approach scored the best in terms of diversity metrics in the ED test dataset.




\subsection{Human Evaluation}

\begin{table*}
\centering
\scriptsize
\begin{tabular}{l | r r r r | r r r r }
\toprule

\multirow{3}{*}{Model} & \multicolumn{4}{c|}{Trained on: OS + ED} & \multicolumn{4}{c}{Trained on: OS + EDOS} \\
& \multicolumn{4}{c|}{Tested on: ED} & \multicolumn{4}{c}{Tested on: EDOS} \\
\cline{2-9}

& Good & Okay & (Good + & Bad & Good & Okay & (Good + & Bad\\
& & & Okay) & & & & Okay) & \\

\midrule


End-to-end model & 46.94 & 15.72 & 62.66 & 37.34 & 27.48 & 27.02 & 54.50 & 45.50\\
Equally sampled & 25.22 & 25.22 & 50.44 & \cellcolor{pink} 49.57 & 19.35 & 29.84 & 49.18 & \cellcolor{pink} 50.82\\
DT (argmax) & 37.61 & \cellcolor{green} 33.70 & \cellcolor{green} 71.31 & 28.69 & 21.70 & \cellcolor{green} 36.72 & \cellcolor{green} 58.42 & 41.58\\
DT (prob. sampled) & 24.17 & 21.68 & 45.85 & 42.13 & 21.96 & 31.78 & 53.74 & 46.26\\
Neural predictor & \cellcolor{green} 50.00 & 17.39 & 67.39 & 32.61 & \cellcolor{green} 24.82 &33.01 & 57.83 & 42.17\\
 
\bottomrule
\end{tabular}
\caption{Human evaluation results (as a percentage) corresponding to ED and EDOS testing datasets.}
\label{tab:human}
\end{table*}

\begin{table*}[ht!]
\scriptsize
\centering
\begin{tabular}{l l l l}
\toprule
\multicolumn{4}{l}{\textbf{Dialogue context:}}\vspace{1mm}\\

\multicolumn{4}{l}{Speaker: \textit{Last week when Toys R Us closed it really made me start thinking of the 90s and my childhood} \textbf{(Nostalgic)}}\\

\multicolumn{4}{l}{Listener: \textit{I've heard a lot of people say something similar. Did you go there a lot?} \textbf{(Questioning)}} \\

\multicolumn{4}{l}{Speaker: \textit{I did! I was talking to my husband about the different toys I had. Even started talking about the music and shows from that era.} \textbf{(Nostalgic)}} \\


\midrule

Model & Predicted emotion/intent  & Generated response & Human eval.\\
\midrule

Ground-truth emotion/intent & Jealous	& \textit{That sounds like a lot of fun. I wish I had a husband like that.} &	Good\\

End-to-end model & -	& \textit{\textcolor{red}{I'm sorry to hear that.} What kind of toys were you talking about?}	& Bad\\

Equally sampled & Wishing &	\textit{\textcolor{red}{I wish you the best of luck!}} &	Bad\\


DT (argmax) & Questioning &	\textit{What kind of toys were you talking about?} &	Good\\


DT (prob. sampled) & Encouraging &	\textit{I bet that was fun} &	Good \\

Neural predictor & Encouraging &	\textit{I bet that was fun} &Good \\

\bottomrule
\end{tabular}
\caption{An example dialogue showing that both lack of controllability and conditioning the response on an inappropriate emotion or intent can lead to responses that are empathetically inappropriate with the dialogue context.}
\label{tab:wrong}
\end{table*}

In addition to the automatic metrics, we carefully designed a human evaluation experiment in Amazon Mechanical Turk (AMT) to evaluate responses' empathetic appropriateness. We selected a total of 1,000 dialogue cases: 500 ED and EDOS dialogues for testing. The AMT workers had to drag and drop responses generated by five models (end-to-end; models whose response was conditioned on the equally sampled baseline, DT argmax, DT prob. sampled and the neural predictor) into areas \textit{Good}, \textit{Okay}, and \textit{Bad}, depending on their empathetic appropriateness. We neglected responses conditioned on the ground-truth emotion or intent since we are more interested in automatically predicted labels. We bundled 10 dialogues into a HIT (Human Intelligence Task) so that one worker works on at least 10 cases to avoid too much bias between answers. To evaluate the quality of the work generated, we included three quiz questions equally spaced in a HIT. In these, we included the ground-truth response among the other responses generated by the models. If a worker rated the ground-truth response either as \textit{Good} or \textit{Okay}, then a bonus point was added. To encourage attentiveness to the task, for those who obtained at least two out of three quiz questions correct, we gave a bonus of 0.1\$. Three workers were allowed to work on a HIT and only the ratings that were agreed by at least two workers, both who have obtained bonuses, were taken to compute the final scores. As a result, 8.33\% of the answers were disqualified. The results of the experiment are denoted in Table \ref{tab:human}. The experiment yielded an inter-rater agreement (Fleiss' kappa) score of $0.2294$ indicating fair agreement.

According to the results, the neural predictor scored the highest percentage of \textit{Good} ratings in both ED and EDOS testing datasets. The models that use the equally sampled approach performed the worst producing the highest percentage of responses ranked \textit{Bad}. An interesting observation is that the DT (argmax) method scored the most number of combined \textit{Good} and \textit{Okay} responses in ED and EDOS testing datasets confirming that rule-baled approaches such as the decision tree approach we propose could be used to control and interpret the responses without losing significant accuracy. 


\subsection{Case Study}

In Table \ref{tab:wrong} we show some example responses generated by different models for a given dialogue context. It could be noticed that having no response control mechanism and having a response conditioned on an inappropriate intent both can result in responses that are empathetically inappropriate with the dialogue context. The neural predictor, as well as the decision tree-based mechanisms, generate some emotion or intent that is appropriate to the dialogue context, enabling the generation model to generate responses that are more empathetically appropriate, guiding the conversation in a meaningful direction.  

\section{Discussion and Conclusion}

This study investigated the use of a taxonomy of empathetic response intents along with 32 fine-grained emotions in controlling and interpreting the responses generated by open-domain conversational agents for emotional prompts. In this regard, several rule-based and automatic response control methods were proposed and were compared in terms of their prediction and generation performance on two state-of-the-art dialogue datasets containing emotional dialogues.

It was observed that the neural response emotion/intent predictor we proposed outperformed the rest including the end-to-end model in terms of evaluation metrics related to both prediction and generation. This implies the importance of leveraging semantic clues in addition to the flow of emotions and intents in the previous turns when predicting the next response's emotion or intent. However, there are some disadvantages to using this approach: 1) developers cannot foresee the label that the model would predict next; and 2) cost of time and resources spent for training the model. As a remedy, we proposed two decision tree-based response emotion/intent prediction approaches. 

Across evaluation metrics for prediction and generation, the performance of the decision-tree methods was considerably better than the end-to-end approach and the equally sampled baseline. The decision tree (argmax) method performed the best in terms of diversity metrics related to response generation. In the human evaluation stage, we saw that the DT (argmax) method produced the most number of combined \textit{Good} and \textit{Okay} responses in ED and EDOS test datasets, pointing to the fact that the rule-based approaches we proposed can still be used without a significant degrade in performance in resource-limited environments. 

On the whole, the results of this study inform developers about the utility of the taxonomy of empathetic response intents in controlling the responses generated by open-domain chatbots and which optimal methodology to use (rule-based or automatic conditioning) based on the operational environment.


\bibliography{acl_latex}
\bibliographystyle{acl_natbib}

\appendix



\appendix

\section{Hyper-parameters used and additional training details}
\label{app:neural}

The BERT-base architecture with 12 layers, 768 dimensions, 12 heads, and 110M parameters is used as the representation network in both the neural response intent predictor and the response generation model. It is initialized with weights from the pre-trained language model RoBERTa \cite{roberta}, which is proven to perform better than pre-trained BERT. We used the same hyper-parameter setting used in RoBERTa (Liu et al., 2019). We used the Adam optimizer with $\beta_1$ of $0.9$, $\beta_2$ of $0.98$, an $\epsilon$ value of $1\times10^{-6}$, and a learning rate of $5\times10^{-5}$. A dropout of 0.1 was used on all layers and attention weights, and a GELU activation function \cite{gelu}. We limited the maximum number of input tokens to $100$, and used a batch size of $256$. All the models were first trained on the OS dialogues dataset for $50$ epochs and fine-tuned on the EmpatheticDialogues and OSED datasets for $10$ epochs each. The best model was chosen based on the minimum loss computed on the validation set after each epoch. The number of training epochs taken for each of the models to converge are denoted in Table \ref{tab:epochs}. All the experiments were conducted on a machine with 2x12cores@2.5GHz, 256 GB RAM, 2x240 GB SSD, and 2xGPU (NVIDIA Titan X Maxwell). 

\begin{table}[ht!]
\small
\centering
\begin{tabularx}{\linewidth}{X X}
\toprule
Model setting & No. of training epochs to converge \\
\midrule
\multicolumn{2}{l}{Neural response emotion and intent predictor}\\
\midrule
OS & 2 epochs (pre-training)\\
OS + ED & 1 epoch (fine-tuning)\\
OS + EDOS & 1 epoch (fine-tuning)\\
\midrule
\multicolumn{2}{l}{Response Generator}\\
\midrule
End-to-end (OS) & 50 epochs (pre-training) \\
End-to-end (OS + ED) & 9 epochs (fine-tuning) \\
End-to-end (OS + EDOS) & 3 epochs (fine-tuning) \\
\midrule
End-to-end + Neural Predictor (OS) &  50 epochs (pre-training)\\
End-to-end + Neural Predictor (OS + ED) & 9 epochs (fine-tuning)\\
End-to-end + Neural Predictor (OS + EDOS) & 3 epochs (fine-tuning)\\
\bottomrule
\end{tabularx}
\caption{The number of training epochs taken for the models to converge}
\label{tab:epochs}
\end{table}

\section{Additional details of the AMT experiment}
\label{app:amt}

Table \ref{tab:amt_stats} indicates the statistics of the Amazon Mechanical Turk experiment conducted to evaluate responses generated by different models. As further measures to control the quality of the workers, we monitored the total time spent on a particular task and reject work that were completed in less than two minutes. And to prevent workers from monopolising the task by accepting a large number of HITs, we warned and blocked the workers who accepted more than 30 HITs. 


\begin{table}[ht!]
\small
\centering
\begin{tabularx}{\linewidth}{X r}
\toprule
Description  & Statistics \\
\hline

No. of HITs & 200\\
& (10 dialogues each)\vspace{1mm}\\
No. of assignments \vspace{1mm}& 400 (200 $\times$ 2)\\
No. of workers \vspace{1mm}& 160\\
Percentage of bonuses earned \vspace{1mm}& 91.67\% \\
No. of blocked workers (Either due to completing more than 30 HITs or due to submitting an assignment in less than 2 minutes \vspace{1mm}& 9 \\
No. of rejected assignments (due to time being < 2 min.s) \vspace{1mm}& 9 \\
Percentage of 2 out of 3 worker agreements on ratings \vspace{1mm}& 88.43\%  \\
Inter rater agreement (Fleiss' & 0.2294\\
kappa) & (fair agreement)\\

\hline
\end{tabularx}
\caption{Statistics of the AMT human evaluation experiment.}
\label{tab:amt_stats}
\end{table}

\begin{figure}[ht!]
\centering
\includegraphics[width=\linewidth]{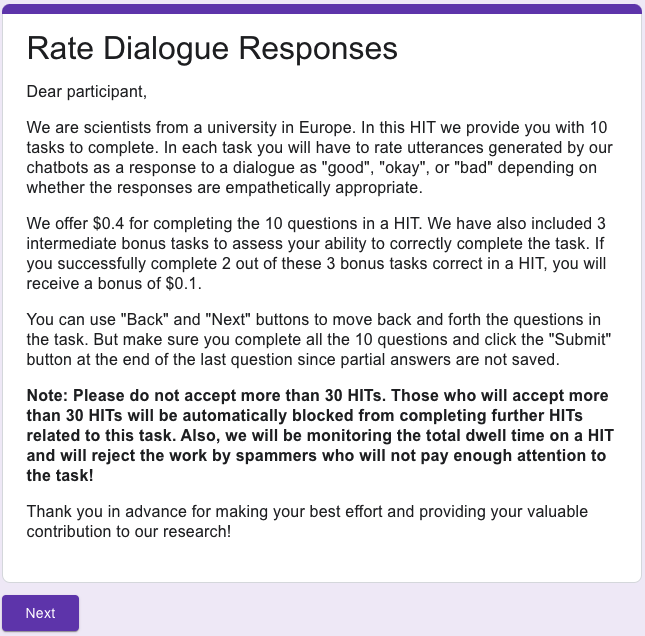} 
\caption{Guidelines of the AMT task for dialogue response evaluation.}
\label{fig:amt-guidelines}
\end{figure}

\begin{figure}[ht!]
\centering
\includegraphics[width=\linewidth]{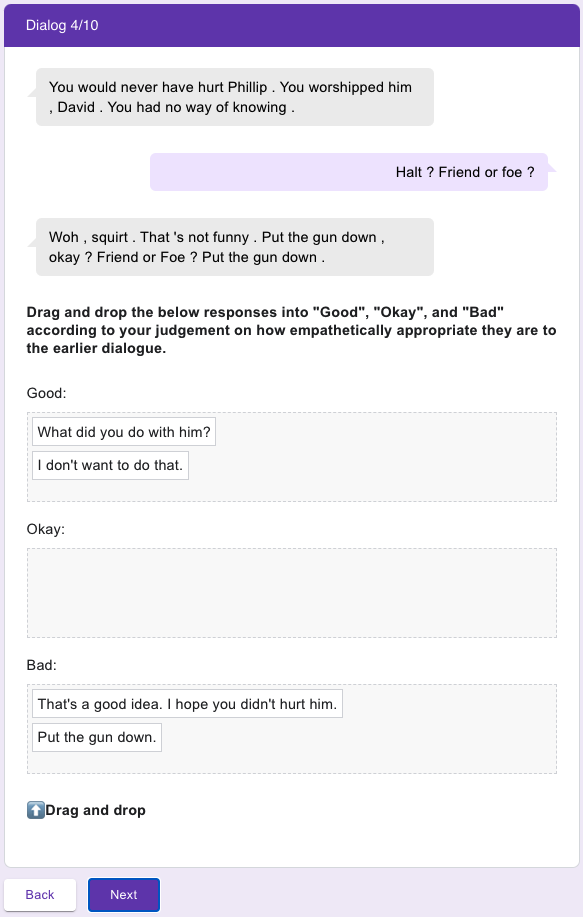} 
\caption{AMT user interface designed for evaluating dialogue responses.}
\label{fig:amt-interface}
\end{figure}

Figures \ref{fig:amt-guidelines} and \ref{fig:amt-interface} show the guidelines provided to the crowd workers and the AMT task interface design, respectively. 


\end{document}